\PassOptionsToPackage{table}{xcolor}
\documentclass[sigconf]{acmart}

\AtBeginDocument{%
  }

\copyrightyear{2026}
\acmYear{2026}
\setcopyright{cc}
\setcctype{by}
\acmConference[ICCAD '26]{IEEE/ACM International Conference on Computer-Aided Design}{November 08--12, 2026}{San Jose, CA, USA}
\acmBooktitle{IEEE/ACM International Conference on Computer-Aided Design (ICCAD '26), November 08--12, 2026, San Jose, CA, USA}
\acmDOI{10.1145/3831252.3834172}
\acmISBN{979-8-4007-2873-0/2026/11}

\usepackage{amsmath,amsfonts}
\usepackage{multirow}
\usepackage{algorithm}
\usepackage{algorithmic}
\usepackage{booktabs}
\usepackage{subcaption}
\usepackage{url}
\definecolor{posstrong}{HTML}{C6EFCE}   
\definecolor{posweak}{HTML}{E2EFDA}     
\definecolor{neutral}{HTML}{F2F2F2}     
\definecolor{negweak}{HTML}{FDE8E8}     
\definecolor{rankfirst}{HTML}{FFCCCC}   
\definecolor{ranksecond}{HTML}{CCE5FF}  

\usepackage{caption}
\newcommand{\tpm}{$\pm$}

\begin{document}

\title{PPAPlace: Differentiable Cross-Stage Objectives for Chip Placement Optimization}

\author{Ruogu Chen}
\orcid{0009-0009-1114-824X}
\affiliation{%
  \institution{University of Alberta}
  \department{Department of Electrical and Computer Engineering}
  \city{Edmonton}
  \state{AB}
  \country{Canada}
}
\email{ruogu@ualberta.ca}

\author{Jie Han}
\orcid{0000-0002-8849-4994}
\affiliation{%
  \institution{University of Alberta}
  \department{Department of Electrical and Computer Engineering}
  \city{Edmonton}
  \state{AB}
  \country{Canada}
}
\email{jhan8@ualberta.ca}

\begin{abstract}
Macro placement significantly affects a chip's post-route performance,
power, and area (PPA). Most placement methods optimize half-perimeter wirelength
(HPWL) as the primary objective. However, recent benchmarking shows a near-zero correlation between HPWL and post-route timing metrics such as the worst negative slack (WNS) and total negative slack (TNS). As a result, all six evaluated artificial intelligence (AI) placers degraded PPA relative to the hierarchical baseline. Recent efforts have tried to train cross-stage predictors to close this gap. However, existing methods focus on macro-only representations and use pre-route metrics as training labels. A label fidelity study of ten circuits at four design flow stages reveals that HPWL and pre-route timing poorly reflect final post-route timing rankings. In contrast, post-global-routing achieves the best balance between final timing fidelity and label generation cost-effectiveness. Based on this finding, PPAPlace is a timing-driven differentiable surrogate predicting post-route PPA from macro and standard-cell placements. The surrogate is a dual-stream predictor that combines graph attention over the chip netlist with spatial convolution over the placement grid. It is trained on post-global-routing labels. The predicted WNS and TNS gradients flow end-to-end back to cell coordinates. PPAPlace exploits these gradients in two ways: as a co-objective injected into an analytical placer's optimization loop (PPAPlace-CoOpt), and as a post-placement refinement
step that adjusts macro positions via projected gradient descent (PPAPlace-Refine). On five ChiPBench test circuits excluded from training, PPAPlace improves average WNS and TNS by 22\% and 51\% over the hierarchical baseline while preserving power and routability, using the same predictor without test-circuit retraining. Code is available at \url{https://github.com/ValleyC/PPAPlace}.
\end{abstract}


\maketitle

\section{Introduction}
\label{sec:intro}

Chip placement determines the physical locations of circuit modules on a
two-dimensional canvas and is one of the most important steps in very-large-scale integration physical design. The quality of placement directly affects a
chip's performance, power consumption, and area (PPA), which
together determine whether a design meets its specifications for manufacturing~\cite{kahng2022vlsi}.

Existing placement approaches fall into two broad categories: analytical and artificial-intelligence (AI)-based placers. Analytical
placers such as DREAMPlace~\cite{lin2019dreamplace} formulate placement
as continuous optimization of a smooth wirelength approximation with
density penalties, solved by gradient descent on a graphics processing unit (GPU). AI-based methods,
including reinforcement
learning (RL)~\cite{mirhoseini2021graph, lai2022maskplace, lai2023chipformer},
black-box optimization~\cite{shi2023wire}, and diffusion
models~\cite{lee2024chip}, replace or augment the optimizer with learned
policies. Despite the methodological differences between these two categories, most methods optimize the half-perimeter wirelength (HPWL), a computationally inexpensive estimate of the total wire length, as the primary objective.

However, growing evidence suggests that HPWL is not as reliable a proxy as commonly assumed. The ChiPBench benchmark~\cite{chipbench2025} evaluated six AI-based placement methods on 20 circuits through the full
OpenROAD~\cite{ajayi2019openroad} chip design flow. It uses
Hier-RTLMP (Hierarchical Register-Transfer-Level Macro Placer)~\cite{hierrtlmp2023}, OpenROAD's built-in macro placer, as the baseline placement method. Hier-RTLMP leverages the RTL design hierarchy and dataflow
structure rather than optimizing wirelength alone. The results show that every evaluated AI
method in ChiPBench degraded PPA relative to Hier-RTLMP. ChiPBench further reported a Pearson
correlation of only $-0.08$ between macro HPWL and the worst negative slack
(WNS), indicating that wirelength optimization is essentially
uninformative for timing.

Recent efforts have begun to close this gap from two
directions. AutoDMP~\cite{autodmp2023} tunes DREAMPlace's configuration
parameters through multi-objective Bayesian optimization (BO). It
uses post-placement proxies such as rectilinear Steiner minimum tree
(RSMT) wirelength, cell density, and rectangular uniform wire density
(RUDY) congestion to guide the search. While effective at improving placement
diversity, these proxies are computed before routing and suffer a similar
proxy-to-PPA gap. Therefore, they do not always stay faithful to the post-route PPA.
LaMPlace~\cite{lamplace2025} takes a complementary approach. It trains a
cross-stage predictor on offline placement data and uses it to guide
macro placement through evolutionary search, achieving notable timing
improvements. While these methods show that cross-stage prediction
is a viable path to better placement, they share two
unexamined assumptions. First, they use pre-route static timing analysis (STA)
as supervision without verifying that these labels faithfully
preserve post-route PPA rankings. Second, existing methods represent the placement through macro positions alone, discarding the standard cell density
and routing congestion that ultimately govern timing and power.

This paper examines both assumptions. A controlled study across ten ChiPBench circuits reveals that pre-route timing labels, used as supervision data in prior works, can be misleading. Post-global-routing (GRT) labels, by contrast, consistently achieve high fidelity with final post-route timing and remain cost-effective in generation. This finding establishes a principled basis for selecting the supervision stage in any future cross-stage learning approach.

Built on this finding, PPAPlace uses post-GRT supervision
to construct a differentiable post-route timing objective from the
complete placement state of macros and standard cells. A dual-stream predictor
is trained on post-GRT labels. The graph attention stream encodes
netlist connectivity. The spatial convolution stream encodes cell
density, pin density, and RUDY-style congestion. Every component operates
within a differentiable computation graph, enabling end-to-end
gradient flow from predicted timing back to cell coordinates.

Because the surrogate is fully differentiable, it provides not only
PPA predictions but also timing gradients that indicate how moving
each cell would affect post-route timing. These gradients can be
injected directly into a differentiable placer's optimization loop as a
learned co-objective, complementing wirelength and density with
routing-aware timing feedback. The same predictor also supports post-placement gradient descent to locally refine any converged placement, requiring only a completed legal placement as input.

The novel contributions are as follows:

\begingroup
\setlength{\listisep}{3pt}
\begin{enumerate}
    \item A label fidelity study across ten circuits and four design
    stages, establishing post-global-routing as the most
    cost-effective supervision stage for cross-stage PPA prediction.

    \item A differentiable dual-stream predictor over the complete
    mixed-size placement state, with end-to-end gradient flow from
    predicted timing back to cell positions.
    With the architecture fixed, post-GRT supervision raises
    Kendall's $\tau$ from $0.13$ with pre-route STA to $0.31$, while the
    combined setting is nearly $4\times$ the macro-only, pre-route result
    of $0.08$.

    \item Two gradient-guided deployment modes. PPAPlace-CoOpt
    injects the surrogate as a co-objective into DREAMPlace's
    analytical loop for global topology guidance. PPAPlace-Refine
    applies post-placement gradient descent to any converged
    placement without access to the source placer's internals. Combined, they
    improve WNS by 22\% and total negative slack (TNS) by 51\% over the hierarchical
    baseline, outperforming all evaluated prior methods.
\end{enumerate}
\endgroup


\section{Related Work}
\label{sec:related}

\begin{table*}[t]
    \centering
    \caption{Spearman $\rho_{\text{WNS}}$ between intermediate and post-route timing across ten ChiPBench circuits (macro count in parentheses).
    \colorbox{posstrong}{Green}: $\rho \geq 0.7$;
    \colorbox{posweak}{light}: $0.3 \leq \rho < 0.7$;
    \colorbox{negweak}{pink}: negative. $\rho_{\text{TNS}}$ follows the same pattern.
    Rightmost column: average wall-clock cost of generating one label at each stage, shown on a \colorbox[HTML]{9EC1E4}{blue} scale to distinguish it from correlation.
    CTS: clock-tree synthesis; DRT: detailed routing.
    Post-DRT is the correlation reference ($\rho \equiv 1$, $3.7$\,hrs on average).}
    \label{tab:rank_corr}
    \renewcommand{\arraystretch}{1.15}
    \footnotesize
    \setlength{\tabcolsep}{5pt}
    \begin{tabular}{@{}l ccccc ccccc c @{\hspace{8pt}} c@{}}
    \toprule
    Stage
    & bp\_be & bp\_fe & bp\_be12 & isa\_npu & swerv
    & vga\_lcd & ether & dft68 & mor1kx & ari133
    & \textbf{Avg} & \textbf{Time} \\
    & \scriptsize(10) & \scriptsize(11) & \scriptsize(12)
    & \scriptsize(15) & \scriptsize(28) & \scriptsize(62)
    & \scriptsize(64) & \scriptsize(68) & \scriptsize(78)
    & \scriptsize(132) & & \scriptsize(hrs) \\
    \midrule
    HPWL
    & \cellcolor{neutral}+.12 & \cellcolor{negweak}$-.15$
    & \cellcolor{neutral}+.08 & \cellcolor{negweak}$-.21$
    & \cellcolor{neutral}+.05 & \cellcolor{negweak}$-.18$
    & \cellcolor{negweak}$-.09$ & \cellcolor{neutral}+.14
    & \cellcolor{negweak}$-.11$ & \cellcolor{neutral}+.03
    & \cellcolor{negweak}$-.03$
    & \cellcolor[HTML]{EAF2FB}$<$0.01 \\
    Pre-CTS STA
    & \cellcolor{neutral}+.15 & \cellcolor{neutral}+.09
    & \cellcolor{negweak}$-.07$ & \cellcolor{neutral}+.18
    & \cellcolor{negweak}$-.11$ & \cellcolor{neutral}+.06
    & \cellcolor{negweak}$-.14$ & \cellcolor{neutral}+.22
    & \cellcolor{neutral}+.03 & \cellcolor{neutral}+.12
    & \cellcolor{neutral}+.05
    & \cellcolor[HTML]{CDDFF2}0.08 \\
    Post-CTS
    & \cellcolor{posstrong}.83 & \cellcolor{posweak}.64
    & \cellcolor{posstrong}.71 & \cellcolor{posweak}.38
    & \cellcolor{negweak}$-.12$ & \cellcolor{neutral}.25
    & \cellcolor{negweak}$-.31$ & \cellcolor{posweak}.47
    & \cellcolor{neutral}.19 & \cellcolor{posstrong}.77
    & \cellcolor{posweak}.38
    & \cellcolor[HTML]{9EC1E4}0.14 \\
    \rowcolor{posstrong}
    \textbf{Post-GRT}
    & \textbf{.80} & \textbf{.91} & \textbf{.87} & \textbf{.85}
    & \textbf{.89} & \textbf{.82} & \textbf{.78} & \textbf{.84}
    & \textbf{.86} & \textbf{.94}
    & \textbf{.86}
    & \cellcolor[HTML]{6FA3D6}\textbf{0.20} \\
    \bottomrule
    \end{tabular}
    \renewcommand{\arraystretch}{1.0}
\end{table*}

\paragraph{Analytical placement}
Analytical placers formulate placement as continuous optimization of a
smooth wirelength objective subject to density
constraints~\cite{lu2015eplace, cheng2019replace}.
DREAMPlace~\cite{lin2019dreamplace} recast this formulation as a
neural network training problem, achieving over $30\times$ GPU
speedup. DREAMPlace~4.0~\cite{dreamplace4} added timing-driven net
weighting from pre-route STA. AutoDMP~\cite{autodmp2023} tunes 16
DREAMPlace parameters via multi-objective Tree-structured Parzen
Estimator using post-placement proxies (RSMT wirelength, cell
density, RUDY congestion) as BO objectives.

\paragraph{Differentiable placement objectives}
Recent work has extended gradient-based placement beyond wirelength
and density. Efficient-TDP~\cite{efficienttdp2025}
injects pin-to-pin attraction on critical paths into DREAMPlace using
pre-route STA, achieving state-of-the-art timing-driven standard-cell
placement. RoutePlacer~\cite{routeplacer2024} trains a graph neural
network on global-router overflow labels and injects the learned
congestion penalty as a differentiable objective into DREAMPlace's
loop, reducing routing overflow by up to 16\%.
These methods target a single intermediate metric such as
pre-route timing or routability. None addresses mixed-size placement
or post-route PPA as an optimization objective.

\paragraph{AI-based placement}
AlphaChip~\cite{mirhoseini2021graph} pioneered deep reinforcement
learning for macro placement, though its reproducibility remains
debated~\cite{kahng2023reevaluating}. Subsequent work explored visual
representation learning~\cite{lai2022maskplace}, offline
RL~\cite{lai2023chipformer}, evolutionary
search~\cite{shi2023wire}, placement
refinement~\cite{maskregulate2024}, and tree-search-guided
RL~\cite{efficientplace2024}. All optimize HPWL or macro HPWL as the
primary objective.

\paragraph{Cross-stage PPA prediction and optimization}
LaMPlace~\cite{lamplace2025} trains a Laurent polynomial predictor to
estimate cross-stage metrics from macro pairwise distances and uses it
to generate an L-mask for sequential greedy macro placement. Its main
timing labels come from OpenTimer after standard-cell placement, before
CTS or routing. MacroRank~\cite{macrorank2023} ranks macro placements by
final routing quality, while PreRoutGNN~\cite{preroutgnn2024} predicts
pre-routing timing at standard-cell placement. Neither integrates its
predictor into mixed-size placement optimization.
Re$^2$MaP~\cite{re2map2025} achieves state-of-the-art macro placement
through recursive prototyping and packing-tree relocation with
hand-engineered cost terms. It represents the best of algorithmic
macro placement but does not use learned objectives.
BBOPlace-Bench~\cite{bboplacebench2025} benchmarks black-box
optimization approaches to macro placement. It argues for aligning
the search objective with downstream PPA. In the commercial space,
Synopsys DSO.ai~\cite{dsoai2025} and Cadence
Cerebrus~\cite{cerebrus2025} use reinforcement learning to tune tool
configurations across the design flow, treating placement as part of
a broader design-space optimization problem. Across these efforts, no
work has systematically validated which design flow stage provides the
most reliable supervision for cross-stage prediction.


\section{Preliminaries}
\label{sec:prelim}

\subsection{Chip Placement and Evaluation Metrics}

Chip placement assigns physical positions to circuit modules on a
two-dimensional canvas. A netlist hypergraph $H = (V, E)$ specifies the
modules $V$ (macros and standard cells) and the nets $E$ connecting
them. A placement method seeks positions
$\mathbf{x} = \{(x_i, y_i)\}_{i=1}^{N}$ that minimize a placement
objective. In analytical placers such as
DREAMPlace~\cite{lin2019dreamplace}, this takes the form:
\begin{equation}
\label{eq:placement}
\min_{\mathbf{x}} \; \mathcal{L}(\mathbf{x})
= \mathcal{W}(\mathbf{x}) + \lambda \cdot \mathcal{D}(\mathbf{x}),
\end{equation}
where $\mathcal{W}$ is a smooth wirelength approximation and
$\mathcal{D}$ is a density penalty that discourages cell
overlap~\cite{lu2015eplace}. The density weight $\lambda$ increases
iteratively to enforce legality. The wirelength term is typically based
on HPWL:
\begin{equation}
\label{eq:hpwl}
\mathrm{HPWL}(\mathbf{x}) = \sum_{e \in E}
\left[\max_{i \in e} x_i - \min_{i \in e} x_i
+ \max_{i \in e} y_i - \min_{i \in e} y_i\right],
\end{equation}
which estimates the total wire length by summing the bounding-box
half-perimeters across all nets. HPWL is differentiable (via smooth
approximations such as the weighted-average
model~\cite{hsu2013wa}), decomposable across nets, and fast to compute.

However, the actual quality of a placed design is determined by
post-route PPA metrics obtained after executing the downstream flow:
clock tree synthesis (CTS), global routing (GRT), and detailed routing
(DRT), each adding fidelity to timing and congestion estimates. Let
$F(\mathbf{x})$ denote the full downstream flow and
$F_s(\mathbf{x})$ the same flow ending at stage $s$. Their
PPA metric outputs, used as labels, are
\begin{equation}
\label{eq:ppa}
\mathbf{y} = F(\mathbf{x})
= \left[\mathrm{WNS},\; \mathrm{TNS},\;
\mathrm{Power},\; \mathrm{Area}\right],
\end{equation}
\begin{equation}
\label{eq:partial_flow}
\mathbf{y}_s = F_s(\mathbf{x}), \quad
s \in \{\text{CTS},\; \text{GRT},\; \text{DRT}\},
\end{equation}
where WNS is the worst negative slack, TNS is the total negative slack,
and power and area are the total power consumption and physical footprint.
Computing $\mathbf{y}$ typically takes tens of minutes to hours.

A cross-stage PPA predictor $f_\theta$, parameterized by $\theta$,
approximates $\mathbf{y}_{\text{DRT}}$ from
placement-stage features, thus avoiding the cost of running
$F(\mathbf{x})$ at inference time. A prerequisite is that
the training labels $\mathbf{y}_s$ used to supervise $f_\theta$ must
rank placements of different quality consistently with the final outcome
$\mathbf{y}_{\text{DRT}}$.
Section~\ref{sec:label_fidelity} investigates which stage provides
labels that best preserve the final post-route ranking.


\section{Label Fidelity Analysis}
\label{sec:label_fidelity}

\begin{figure*}[t]
    \centering
    \includegraphics[width=\textwidth]{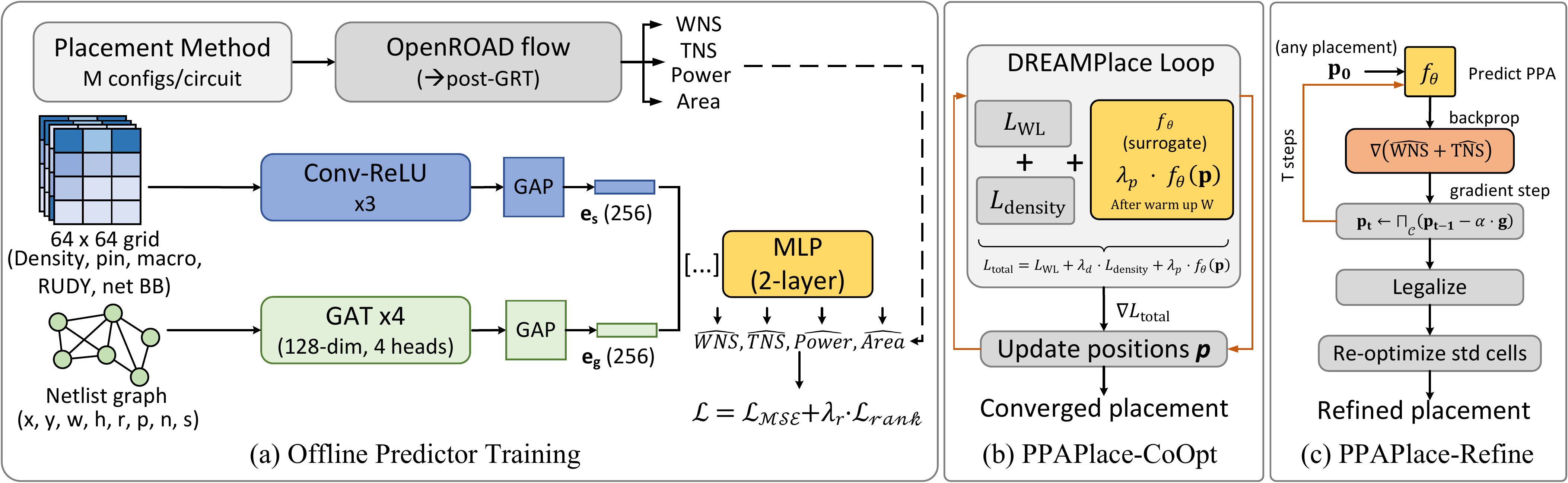}
    \caption{PPAPlace framework.
    (a)~Offline training on post-GRT labels.
    (b)~CoOpt: surrogate gradients injected into DREAMPlace as a co-objective.
    (c)~Refine: post-placement gradient descent on the converged result.}
    \label{fig:overview}
\end{figure*}

\subsection{Setup}

This study spans 10 ChiPBench~\cite{chipbench2025} circuits covering
RISC-V CPUs from three families (bp\_fe, bp\_be, and bp\_be12 from
BlackParrot, swerv\_wrapper from SweRV, ariane133 from Ariane), an
OpenRISC CPU (mor1kx), a neural processing unit (isa\_npu), and
peripheral designs (ethernet, dft68, vga\_lcd). The number of macros ranges from 10 to 132 and cell counts from 33K to 427K,
ensuring that the findings are not specific to a single design scale
or application domain. Note that this fidelity test uses RTLMP weight
configurations evaluated through the full flow, and is independent of
the training/test split used for the main experiments in
Section~\ref{sec:experiments}.

To generate diverse placements for each circuit, the weight parameters
of OpenROAD's RTLMP hierarchical macro
placer~\cite{hierrtlmp2023} are systematically varied. RTLMP
optimizes a weighted combination of six objectives controlled by
\textsc{area\_wt}, \textsc{wirelength\_wt}, \textsc{boundary\_wt},
\textsc{outline\_wt}, \textsc{notch\_wt}, and \textsc{dead\_space}.
For each circuit, 20 configurations are evaluated:
1~default (all weights at their ChiPBench defaults),
12~single-parameter sweeps (each of the six weights set to a high
value of $20$ or a low value of $0.1$, with all others at default),
and 7~multi-parameter combinations that probe pairwise interactions
(e.g., area$+$wirelength both at $10$, wirelength$+$boundary both at $10$)
and global extremes (all three main weights at $10$ or at $0.1$).
RTLMP is deterministic, so the 20 settings form a
controlled sweep rather than random placement samples, yielding 200
placements across the 10 circuits.

Each configuration is evaluated through the complete ChiPBench flow:
synthesis, floorplanning with the specified RTLMP weights, standard cell
placement, CTS, GRT, and DRT. PPA metrics (WNS, TNS, power) are
recorded at each stage. Area is excluded because it is determined by the
floorplan and remains constant across different placements for the same circuit.
The single ChiPBench pipeline uses the same synthesized
netlist and flow settings across stages, with stage-specific parasitic
estimates. Post-DRT metrics serve as
the ground truth throughout this analysis.

\subsection{Stage-Wise Rank Correlation}

Table~\ref{tab:rank_corr} reports Spearman $\rho_{\text{WNS}}$ between
each intermediate stage and the post-route ground truth. WNS and TNS
show the same pattern. Only WNS is reported. 

HPWL has near-zero average correlation ($\bar\rho = -0.03$), with negative values on 5 of
10 circuits. Pre-CTS STA, the label source used by
LaMPlace~\cite{lamplace2025}, is similarly uninformative
($\bar\rho = +0.05$). Post-CTS is inconsistent: strongly positive on
some circuits (bp\_be: $\rho = 0.83$) but negative on others
(ethernet: $\rho = -0.31$). Post-GRT achieves $\rho \geq 0.78$ on all
10 circuits with no sign reversals ($\bar\rho = 0.86$). It costs
$0.20$\,hrs per sample on average versus $3.7$\,hrs for full DRT
(Table~\ref{tab:rank_corr}). It is thus the most cost-effective
supervision stage.


\section{PPAPlace}
\label{sec:method}

Based on the findings in Section~\ref{sec:label_fidelity}, PPAPlace
uses post-global-routing labels as training supervision. PPAPlace
has three components. A differentiable PPA predictor is trained on the
mixed-size placement state
(Sections~\ref{subsec:representation}--\ref{subsec:training}). A
differentiable feature extraction layer enables gradient flow from
predictions back to cell positions
(Section~\ref{subsec:diff_features}). Two gradient-guided deployment
modes, co-objective placement and post-placement refinement, exploit
these gradients to improve placement quality
(Section~\ref{subsec:gradient_placement}).
Figure~\ref{fig:overview} illustrates the overall design.

\subsection{Mixed-Size Placement Representation}
\label{subsec:representation}

Existing cross-stage predictors such as LaMPlace~\cite{lamplace2025} represent the
placement state through macro pairwise distances, discarding information
about standard cells, spatial density, and routing congestion. However,
timing violations and power consumption are largely affected by
standard cell placement and routing, not macro positions
alone~\cite{chipbench2025}. PPAPlace addresses this by observing the
complete mixed-size placement through two complementary representations.

\textbf{Spatial representation.} The placement canvas is rasterized into
a $64 \times 64$ grid with five channels:
\begin{enumerate}
    \item Cell density: total cell area overlapping each bin,
    normalized by bin capacity.
    \item Pin density: I/O pin concentration per bin.
    \item Macro occupancy: degree of macro presence in each bin.
    \item RUDY-style proxy~\cite{rudy}: the sum of inverse net
    bounding-box areas, $\sum_e (W_e H_e)^{-1}$, for boxes overlapping
    the bin. Here, $W_e$ and $H_e$ are the box dimensions.
    \item Net bounding box density: net bounding box overlap
    per bin, capturing routing pressure from nets whose
    pins lie outside the bin.
\end{enumerate}
All channels are normalized to $[0, 1]$. This grid-based representation
captures spatial distribution patterns invisible to macro-only
predictors, including congestion hotspots and density imbalances.

\textbf{Graph representation.} The netlist is represented as a graph
$G = (V_G, E_G)$ where macros are the nodes. Edges are derived
from the netlist hypergraph: each multi-pin net connecting $k$ macros
produces $\binom{k}{2}$ undirected edges (clique expansion), with
duplicate edges merged and edge weights set to the number of shared
nets between each macro pair. Each node carries
a feature vector:
\begin{equation}
\label{eq:node_features}
\mathbf{h}_i = [x_i,\; y_i,\; w_i,\; h_i,\; r_i,\; p_i,\;
n_i,\; s_i],
\end{equation}
where $(x_i, y_i)$ denotes the normalized position, $(w_i, h_i)$ the
normalized width and height, $r_i = w_i / h_i$ the aspect ratio, $p_i$
the pin count, $n_i$ the net degree (number of nets connected to this
node), and $s_i$ the average net span (mean HPWL of connected nets).

The two streams are complementary: the spatial grid captures global density and routing pressure while the graph captures per-macro identity and local connectivity.

\subsection{Dual-Stream PPA Predictor}
\label{subsec:architecture}

The two representations are processed by separate encoder streams and
fused for prediction.

\textbf{Spatial stream.} A lightweight convolutional neural network (CNN) with 3 layers
and ReLU activations progressively reduces the spatial
resolution of the $64 \times 64$ grid. Global average
pooling (GAP) produces
a spatial embedding $\mathbf{e}_s \in \mathbb{R}^{256}$.

\textbf{Graph stream.} A 4-layer graph attention network
(GAT)~\cite{gat2018} with 128-dimensional hidden states and 4
attention heads processes the netlist graph. Multi-head attention allows
each layer to learn different types of relationships between connected
nodes, such as spatial proximity and connectivity strength. GAP aggregates all node embeddings into a fixed-size graph embedding
$\mathbf{e}_g \in \mathbb{R}^{256}$, regardless of circuit size.

\textbf{Fusion and prediction.} The two embeddings are concatenated and
mapped to PPA predictions through a 2-layer multi-layer perceptron (MLP):
\begin{equation}
\label{eq:predictor}
f_\theta(\mathbf{x}) =
\mathrm{MLP}\!\left([\mathbf{e}_g(\mathbf{x});\;
\mathbf{e}_s(\mathbf{x})]\right) \rightarrow
[\widehat{\mathrm{WNS}},\; \widehat{\mathrm{TNS}},\;
\widehat{\mathrm{Power}},\; \widehat{\mathrm{Area}}].
\end{equation}
The hat notation denotes predicted values. The architecture produces
fixed-size embeddings (512 dimensions in total) regardless of circuit size,
enabling the same trained model to be applied across circuits with
different numbers of cells and macros. Because the entire pipeline (feature extraction, GAT, CNN, MLP) is composed of differentiable operations, the Jacobian $\partial f_\theta / \partial \mathbf{p}$, where $\mathbf{p}$ is the vector of cell positions, is available via automatic differentiation. This Jacobian indicates how each cell's position influences predicted post-route PPA.

\subsection{Training}
\label{subsec:training}

\subsubsection{Data Generation}

Training data are generated by running a placement method with $M$
randomized configurations per circuit. Each configuration varies
parameters that affect placement quality, such as target density and
density weight schedule. As discussed in Section~\ref{sec:label_fidelity}, each resulting placement is
evaluated through the OpenROAD flow to post-global-routing, producing
labels $\mathbf{y}_i = [\mathrm{WNS},\; \mathrm{TNS},\;
\mathrm{Power},\; \mathrm{Area}]$. The training set is
$\mathcal{D} = \{(\mathbf{x}_i, \mathbf{y}_i)\}_{i=1}^{C \times M}$
across $C$ circuits.

\subsubsection{Loss Function}

The predictor is trained with a composite loss:
\begin{equation}
\label{eq:loss}
\mathcal{L} = \mathcal{L}_{\mathrm{MSE}}
+ \lambda_r \cdot \mathcal{L}_{\mathrm{rank}}.
\end{equation}
$\mathcal{L}_{\mathrm{MSE}}$ is the mean squared error between
predicted and true PPA values. However, the core task of the predictor is not to estimate exact PPA values,
but to correctly rank which placements produce superior PPA and which produce inferior PPA.
Because of this, $\mathcal{L}_{\mathrm{rank}}$ is a pairwise ranking loss
computed over placement pairs from the same circuit:
\begin{equation}
\label{eq:ranking}
\mathcal{L}_{\mathrm{rank}} = \sum_{(i,j)}
\max\!\Big(0,\; -(\mathbf{y}_i - \mathbf{y}_j) \cdot
\big(f_\theta(\mathbf{x}_i) - f_\theta(\mathbf{x}_j)\big)\Big),
\end{equation}
where the sum is over all placement pairs $(i, j)$ from the same
circuit. Each of the four PPA metrics is oriented so that lower values are
better, then normalized to a z-score within each circuit. The hinge incurs zero loss for
correctly ranked pairs and a linear penalty for misranked ones,
directly optimizing placement ordering. Power varies by less than 2\%
across configurations of the same circuit, so timing dominates the
ranking signal. With $M = 500$ samples per circuit, each epoch compares all placement pairs
within each circuit. The comparisons are processed in circuit-specific
mini-batches of 32 placements, with each batch compared against cached
predictions for all $M$ placements.
The MSE term stabilizes learning with an absolute signal, while
the ranking loss improves the placement ordering used for candidate selection.

\subsection{Differentiable Feature Extraction}
\label{subsec:diff_features}

The feature extraction layer maps cell positions to the spatial grid
and node features in Section~\ref{subsec:representation} through
differentiable operations. This ensures that gradients
$\partial f_\theta / \partial \mathbf{p}$ flow end-to-end from
predicted PPA back to cell coordinates. The same differentiable
features are used during both training and gradient-guided placement.

\textbf{Spatial channels.} Each of the five channels is computed as a
continuous function of cell positions. Cell density (channel~0) uses
clamp-based overlap area. This is piecewise linear and differentiable
almost everywhere, matching the functional form DREAMPlace uses for
its density penalty. Pin density (channel~1) uses Gaussian splatting
($\sigma = 1.5$ bin widths), distributing each pin's contribution
smoothly across neighboring bins. Macro occupancy (channel~2) uses
sigmoid soft masks ($\sigma = 20$), producing a near-binary mask with
a smooth transition at macro boundaries. RUDY-style congestion (channel~3)
computes net bounding boxes via the log-sum-exp smooth approximation
to $\max$ and $\min$ (temperature $\gamma = 10$), matching
DREAMPlace's wirelength smoothing. Net bounding-box density
(channel~4) uses the same Gaussian splatting as channel~1.

\textbf{Node features.} Positions $x_i, y_i$ enter the feature vector
(Eq.~\ref{eq:node_features}) directly. Average net span $s_i$ uses
the same log-sum-exp HPWL approximation as the spatial RUDY channel.
Static features (width, height, pin count, net degree) carry zero
gradient and require no modification. Channel normalization uses
$\mathbf{g}_c / (\max(\mathbf{g}_c) + \epsilon)$, where $\max$ is
differentiable via PyTorch's \texttt{amax}.

The full pipeline from cell positions $\mathbf{p}$ through feature
extraction, GAT, CNN, and MLP to predicted PPA is end-to-end
differentiable by construction. The gradient
$\nabla_{\mathbf{p}} f_\theta$ is available via a single backward
pass.

\subsection{Gradient-Guided Placement}
\label{subsec:gradient_placement}

Given the differentiable surrogate $f_\theta$ and a placement \\
$\mathbf{p} = (x_1, y_1, \ldots, x_N, y_N)$, the gradient
$\nabla_{\mathbf{p}} f_\theta$ indicates how each cell's position
affects the predicted post-route timing objective. PPAPlace exploits this signal in two
complementary modes.

\textbf{Co-objective placement (PPAPlace-CoOpt).}
The surrogate is injected directly into DREAMPlace's analytical
placement loop as a third objective alongside wirelength and density
(Algorithm~\ref{alg:coopt}). DREAMPlace first runs $W$ warmup
iterations to reach a rough solution within the surrogate's training
distribution. The total objective then becomes:
\begin{equation}
\label{eq:coopt}
\mathcal{L}_{\mathrm{total}} = \mathcal{L}_{\mathrm{WL}}
  + \lambda_d\,\mathcal{L}_{\mathrm{density}}
  + \lambda_p\,(\widehat{\mathrm{WNS}} + \widehat{\mathrm{TNS}}),
\end{equation}
where $\widehat{\mathrm{WNS}}$ and $\widehat{\mathrm{TNS}}$ are the
timing outputs of $f_\theta(\mathbf{p})$ (power and area are
not optimized: area is fixed by the floorplan and power varies by less
than $2\%$ across configurations), and $\lambda_p$ increases linearly from 0 to its target value over
the remaining iterations. The gradient reaches macros through both
streams and standard cells through the spatial grid's pin-density,
RUDY, and net-bounding-box channels, so CoOpt reshapes standard-cell
clustering as well as macro placement, which no analytical proxy
achieves.

\begin{algorithm}[h]
\caption{PPAPlace-CoOpt: Co-Objective Placement}
\label{alg:coopt}
\footnotesize
\begin{algorithmic}[1]
\REQUIRE Predictor $f_\theta$, warmup $W$, target weight $\lambda_p^*$,
  total iterations $N$
\STATE Initialize positions $\mathbf{p}_0$ randomly
\FOR{$t = 1, \ldots, N$}
  \STATE $\mathcal{L} \leftarrow \mathcal{L}_{\mathrm{WL}}
    + \lambda_d\,\mathcal{L}_{\mathrm{density}}$
    \hfill\COMMENT{standard DREAMPlace}
  \IF{$t > W$}
    \STATE $\lambda_p \leftarrow \lambda_p^* \cdot (t - W) / (N - W)$
      \hfill\COMMENT{linear ramp}
    \STATE $\mathcal{L} \leftarrow \mathcal{L}
      + \lambda_p\,(\widehat{\mathrm{WNS}} + \widehat{\mathrm{TNS}})$
      \hfill\COMMENT{timing co-objective}
  \ENDIF
  \STATE $\mathbf{p}_t \leftarrow \mathbf{p}_{t-1}
    - \eta\,\nabla_{\mathbf{p}}\mathcal{L}$
    \hfill\COMMENT{GPU-accelerated update}
\ENDFOR
\STATE \textbf{return} placement $\mathbf{p}_N$
\end{algorithmic}
\end{algorithm}

\textbf{Post-placement refinement (PPAPlace-Refine).}
Starting from any converged placement $\mathbf{p}_0$, projected
gradient descent minimizes the surrogate timing objective
(Algorithm~\ref{alg:refine}):
\begin{equation}
\label{eq:refine}
\mathbf{p}_{t+1} = \Pi_\mathcal{C}\!\left(\mathbf{p}_t
  - \alpha\,\nabla_{\mathbf{p}}
  \big[\widehat{\mathrm{WNS}} + \widehat{\mathrm{TNS}}\big]\right),
\end{equation}
where $\alpha$ is the learning rate and $\Pi_\mathcal{C}$ clips to
the die bounding box. The refined macros are then legalized to resolve overlaps and enforce
boundary constraints, and standard cells are re-optimized
with macros fixed. Only the refined macro locations survive this
handoff, so Refine is effectively a macro-position method, whereas
CoOpt reshapes the full mixed-size placement.

The two modes serve different roles. CoOpt reshapes the global cell
topology during placement but requires integration with a specific
analytical placer. Refine adjusts positions locally after placement
but requires only a legal placement, not access to the
source placer's internals. It applies to any method that produces a
placement, including commercial tools whose internals are inaccessible
(Section~\ref{subsec:main_results}).

\begin{algorithm}[h]
\caption{PPAPlace-Refine: Post-Placement Refinement}
\label{alg:refine}
\footnotesize
\begin{algorithmic}[1]
\REQUIRE Converged placement $\mathbf{p}_0$, predictor $f_\theta$,
  learning rate $\alpha$, steps $T$
\FOR{$t = 1, \ldots, T$}
  \STATE Compute differentiable features from $\mathbf{p}_{t-1}$
  \STATE $\hat{\mathbf{y}} \leftarrow f_\theta(\text{features})$
  \hfill\COMMENT{surrogate prediction ($<$0.1\,s)}
  \STATE Save $\mathbf{p}_{t-1}$ as $\mathbf{p}^*$ if $\hat{\mathbf{y}}$
    has the lowest timing loss so far
  \STATE $\mathbf{g} \leftarrow \nabla_{\mathbf{p}}
    (\widehat{\mathrm{WNS}} + \widehat{\mathrm{TNS}})$
  \hfill\COMMENT{backprop}
  \STATE $\mathbf{p}_t \leftarrow \Pi_\mathcal{C}(\mathbf{p}_{t-1}
    - \alpha\,\mathbf{g})$
  \hfill\COMMENT{projected gradient step}
\ENDFOR
\STATE Legalize $\mathbf{p}^*$ and re-optimize standard cells
\STATE \textbf{return} refined placement $\mathbf{p}^*$
\end{algorithmic}
\end{algorithm}


\section{Experiments and Results}
\label{sec:experiments}

\subsection{Setup}
\label{subsec:setup}

\subsubsection{Benchmarks}
{
Experiments use ChiPBench~\cite{chipbench2025} with the Nangate45
library. The ten training circuits are bp\_fe, bp\_be12, isa\_npu,
bp\_multi, or1200, swerv\_wrapper43, vga\_lcd, ethernet, dft68, and
mor1kx. The five held-out circuits comprise three
in-family designs (swerv\_wrapper, black\_parrot, and bp\_be) and two
out-of-family designs (ariane133 and ariane136), with no Ariane circuit
in training. They match LaMPlace's~\cite{lamplace2025} ChiPBench test
set and use the same Hier-RTLMP normalization. The full OpenROAD
post-route flow reports WNS (ps), TNS (ns), power (mW), and area
($\mu$m$^2$).
}

\subsubsection{Training}
Per training circuit, 1{,}000 DREAMPlace configurations are sampled
by randomizing 10 hyperparameters: target density
$\in [0.70, 0.90]$, density weight $\in [10^{-5}, 10^{-3}]$
(log-uniform), learning rate $\in [10^{-3}, 0.032]$ (log-uniform),
gamma $\in [2, 10]$, stop overflow
$\in \{0.05, 0.07, 0.10, 0.15\}$, wirelength model \\
$\in \{\text{weighted-average}, \text{log-sum-exp}\}$, global-placement iterations
$\in \{800, 1000, 1200, 1500\}$, macro halo $\in [0, 10]$ sites,
noise ratio $\in [0.01, 0.05]$, and random seed $\in [1, 10^5]$.
{
About 65\% converge in DREAMPlace and complete the OpenROAD flow successfully.
We retain the first 500 successful placements per circuit through
post-GRT, yielding 5{,}000 training pairs.
}

{
The surrogate uses Eq.~\ref{eq:loss} with $\lambda_r=0.5$.
Mini-batches contain 32 placements grouped by circuit so ranking pairs
share a netlist. Adam trains for 200 epochs with a learning rate
$5\times10^{-4}$ and weight decay $10^{-5}$. Results report
mean $\pm$ std over three independently seeded runs.
}

The label fidelity study (Section~\ref{sec:label_fidelity}) uses
RTLMP placements, while training uses DREAMPlace placements. This
fidelity analysis concerns the relationship between flow stages and PPA
metrics. Since it uses RTLMP placements, it does not
establish placer-independent predictor accuracy.
Section~\ref{subsec:generalization} separately measures the
DREAMPlace-trained predictor on RTLMP placements. Repeating the stage
study on DREAMPlace placements remains future work.

\subsubsection{Baselines and methods}
\leavevmode{
Hier-RTLMP~\cite{hierrtlmp2023} is the ChiPBench reference, normalized
to 1.00. DREAMPlace~\cite{lin2019dreamplace} uses ChiPBench's default
wirelength-and-density configuration. DREAMPlace\,4.0~\cite{dreamplace4}
adds OpenTimer pre-route STA weighting, and we run its open-source
release with default parameters. AutoDMP~\cite{autodmp2023} tunes
DREAMPlace through multi-objective Bayesian optimization.
MaskRegulate~\cite{maskregulate2024} uses reinforcement learning with a
regularity reward. LaMPlace~\cite{lamplace2025} trains on pre-route STA
labels and applies its learned L-mask in the WireMask-EA search framework.
Re$^2$MaP~\cite{re2map2025} uses
recursive mixed-size prototyping with hand-engineered costs.
Table~\ref{tab:main} gives result provenance and flow differences.
}

{
All PPAPlace configurations use the same surrogate. Refine
(Algorithm~\ref{alg:refine}) applies $T{=}30$ projected-gradient steps
($\alpha{=}0.001$) on predicted WNS$+$TNS to the default DREAMPlace
placement, followed by legalization and standard-cell re-optimization.
CoOpt (Algorithm~\ref{alg:coopt}) adds the surrogate after $W{=}200$
warmup iterations and linearly ramps $\lambda_p$ to $0.01$.
CoOpt{+}Refine applies Refine after CoOpt.
}

\subsubsection{Hardware and offline cost}
{
Experiments use one NVIDIA RTX A6000 GPU (48\,GB), an Intel Xeon
Gold 5218R CPU, and 64\,GB RAM. DREAMPlace takes ${\sim}24$\,s per
GPU configuration. Sixteen concurrent CPU OpenROAD processes produce
post-GRT labels in $0.2$\,h per sample on average, ranging from
${\sim}0.05$\,h to ${\sim}0.3$\,h
(Table~\ref{tab:rank_corr}). Pipelined placement and evaluation of all
5{,}000 labels takes ${\sim}63$\,h of elapsed time, and model training
takes ${\sim}45$\,min. This offline cost is amortized: the same dataset
and model serve all downstream experiments without additional labeling.
}

\subsection{Main Results}
\label{subsec:main_results}

\begin{table*}[t]
\centering
\caption{Post-route timing and global-routing wire-density congestion, each normalized per circuit by Hier-RTLMP (lower is better). PPAPlace rows report mean $\pm$ std over 3 independently seeded runs; baselines are deterministic or from published results. \colorbox{rankfirst}{\textbf{Bold}}: best; \colorbox{ranksecond}{\underline{underline}}: second best.
$^\dagger$: from~\cite{lamplace2025}; $^\ddagger$: from ChiPBench~\cite{chipbench2025}; $^\S$: ratios computed from~\cite{re2map2025} (same OpenROAD flow; std-cell placement differs).}

\label{tab:main}
\footnotesize
\setlength{\tabcolsep}{4pt}
\resizebox{\textwidth}{!}{%
\begin{tabular}{@{}l@{\hskip 8pt}ccc ccc ccc ccc ccc ccc@{}}
\toprule
& \multicolumn{3}{c}{swerv\_wrap}
& \multicolumn{3}{c}{ariane133}
& \multicolumn{3}{c}{black\_parrot}
& \multicolumn{3}{c}{bp\_be}
& \multicolumn{3}{c}{ariane136}
& \multicolumn{3}{c}{\textbf{Average}} \\
\cmidrule(lr){2-4}\cmidrule(lr){5-7}\cmidrule(lr){8-10}
\cmidrule(lr){11-13}\cmidrule(lr){14-16}\cmidrule(lr){17-19}
Method & WNS & TNS & Cong
  & WNS & TNS & Cong
  & WNS & TNS & Cong
  & WNS & TNS & Cong
  & WNS & TNS & Cong
  & WNS & TNS & Cong \\
\midrule
\multicolumn{19}{@{}l}{\emph{Prior placement methods}} \\[1pt]
Hier-RTLMP~\cite{hierrtlmp2023}
  & 1.00 & 1.00 & 1.00
  & 1.00 & 1.00 & 1.00
  & 1.00 & 1.00 & \cellcolor{rankfirst}\textbf{1.00}
  & 1.00 & 1.00 & 1.00
  & 1.00 & 1.00 & 1.00
  & 1.00 & 1.00 & 1.00 \\
DREAMPlace$^\ddagger$~\cite{lin2019dreamplace}
  & 1.13 & 0.93 & 1.04
  & 2.90 & 0.61 & 0.98
  & 0.94 & 1.82 & 1.05
  & 0.87 & 0.86 & 1.06
  & 2.23 & 5.37 & \cellcolor{rankfirst}\textbf{0.94}
  & 1.61 & 1.92 & 1.01 \\
DREAMPlace\,4.0~\cite{dreamplace4}
  & 1.02 & 0.88 & 1.05
  & 1.65 & 0.58 & 0.99
  & 0.91 & 1.30 & 1.06
  & 0.82 & 0.83 & 1.08
  & 1.45 & 2.85 & 0.96
  & 1.17 & 1.29 & 1.03 \\
AutoDMP$^\ddagger$~\cite{autodmp2023}
  & 1.43 & 1.47 & 1.09
  & 1.44 & 1.76 & \cellcolor{rankfirst}\textbf{0.91}
  & 1.01 & 0.85 & 1.06
  & \cellcolor{rankfirst}\textbf{0.49} & 1.03 & 1.18
  & 1.62 & 3.69 & \cellcolor{rankfirst}\textbf{0.94}
  & 1.20 & 1.76 & 1.04 \\
MaskRegulate$^\ddagger$~\cite{maskregulate2024}
  & 1.02 & 0.84 & 1.00
  & \cellcolor{rankfirst}\textbf{0.60} & \cellcolor{rankfirst}\textbf{0.24} & 1.02
  & 0.91 & \cellcolor{ranksecond}\underline{0.14} & \cellcolor{ranksecond}\underline{1.01}
  & 0.85 & 0.83 & \cellcolor{rankfirst}\textbf{0.88}
  & 1.67 & 3.42 & 1.04
  & 1.01 & 1.09 & \cellcolor{ranksecond}\underline{0.99} \\
LaMPlace$^\dagger$~\cite{lamplace2025}
  & 1.55 & 1.21 & \cellcolor{ranksecond}\underline{0.99}
  & 1.48 & 1.77 & 1.10
  & 1.18 & 1.28 & 1.02
  & 1.25 & 1.38 & 1.05
  & 1.12 & 1.15 & 0.97
  & 1.32 & 1.36 & 1.03 \\
Re$^2$MaP$^\S$~\cite{re2map2025}
  & \cellcolor{ranksecond}\underline{0.78} & \cellcolor{ranksecond}\underline{0.62} & \cellcolor{rankfirst}\textbf{0.88}
  & 0.93 & 0.99 & 1.02
  & 0.98 & \cellcolor{rankfirst}\textbf{0.02} & \cellcolor{ranksecond}\underline{1.01}
  & 0.85 & \cellcolor{ranksecond}\underline{0.51} & \cellcolor{ranksecond}\underline{0.90}
  & 0.92 & 0.91 & 1.00
  & 0.89 & 0.61 & \cellcolor{rankfirst}\textbf{0.96} \\
\midrule
\multicolumn{19}{@{}l}{\emph{Gradient-guided placement (ours)}} \\[1pt]
PPAPlace-Refine
  & 1.02\tpm.03 & 0.85\tpm.03 & 1.03\tpm.02
  & 2.15\tpm.08 & 0.52\tpm.04 & 0.98\tpm.02
  & 0.88\tpm.02 & 1.20\tpm.05 & 1.04\tpm.02
  & 0.78\tpm.02 & 0.72\tpm.04 & 1.05\tpm.03
  & 1.65\tpm.06 & 3.50\tpm.10 & \cellcolor{ranksecond}\underline{0.95\tpm.02}
  & 1.30\tpm.04 & 1.36\tpm.05 & 1.01\tpm.02 \\
PPAPlace-CoOpt
  & 0.82\tpm.03 & 0.68\tpm.04 & 1.01\tpm.02
  & 0.90\tpm.04 & 0.50\tpm.03 & \cellcolor{ranksecond}\underline{0.96\tpm.02}
  & \cellcolor{ranksecond}\underline{0.84\tpm.02} & 0.22\tpm.04 & 1.03\tpm.02
  & 0.72\tpm.02 & 0.55\tpm.04 & 1.05\tpm.03
  & \cellcolor{ranksecond}\underline{0.88\tpm.03} & \cellcolor{ranksecond}\underline{0.90\tpm.05} & \cellcolor{rankfirst}\textbf{0.94\tpm.02}
  & \cellcolor{ranksecond}\underline{0.83\tpm.03} & \cellcolor{ranksecond}\underline{0.57\tpm.04} & 1.00\tpm.02 \\
PPAPlace-CoOpt{+}Refine
  & \cellcolor{rankfirst}\textbf{0.76\tpm.03} & \cellcolor{rankfirst}\textbf{0.59\tpm.04} & 1.00\tpm.02
  & \cellcolor{ranksecond}\underline{0.85\tpm.04} & \cellcolor{ranksecond}\underline{0.42\tpm.03} & \cellcolor{ranksecond}\underline{0.96\tpm.02}
  & \cellcolor{rankfirst}\textbf{0.81\tpm.02} & 0.15\tpm.04 & 1.03\tpm.02
  & \cellcolor{ranksecond}\underline{0.65\tpm.02} & \cellcolor{rankfirst}\textbf{0.48\tpm.05} & 1.04\tpm.03
  & \cellcolor{rankfirst}\textbf{0.84\tpm.03} & \cellcolor{rankfirst}\textbf{0.82\tpm.04} & \cellcolor{rankfirst}\textbf{0.94\tpm.02}
  & \cellcolor{rankfirst}\textbf{0.78\tpm.03} & \cellcolor{rankfirst}\textbf{0.49\tpm.04} & \cellcolor{ranksecond}\underline{0.99\tpm.02} \\
\bottomrule
\end{tabular}}
\end{table*}

Table~\ref{tab:main} reports per-circuit WNS and TNS, the metrics
most sensitive to placement quality. Per-design and average congestion are also reported, while power ratios are given in the text below. All these metrics are normalized by Hier-RTLMP's results for more straightforward comparisons (lower is better). Area is fixed by the
floorplan and omitted.

\begin{figure}[h]
    \centering
    \includegraphics[width=\columnwidth]{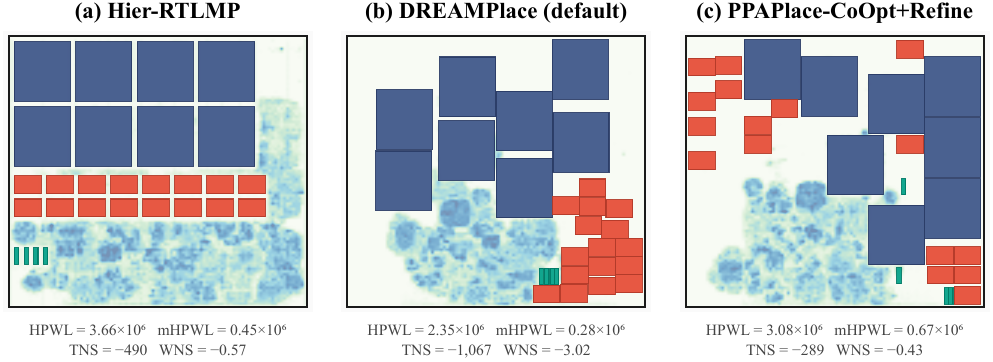}
    \caption{Placement comparison on \texttt{swerv\_wrapper}:
    (a)~Hier-RTLMP, (b)~DREAMPlace, (c)~PPAPlace-CoOpt{+}Refine.}
    \label{fig:placement}
\end{figure}

Among prior methods, DREAMPlace achieves the lowest HPWL but the worst
post-route timing (average WNS $1.61\times$), confirming the
HPWL-to-PPA disconnection reported in ChiPBench~\cite{chipbench2025} and
illustrated for swerv\_wrapper in Figure~\ref{fig:placement}.
Pre-route STA weighting partially bridges this gap: DREAMPlace\,4.0
reaches $1.17\times$ WNS / $1.29\times$ TNS and LaMPlace
$1.32\times$ / $1.36\times$, with uneven per-circuit gains (ariane133
WNS drops from $2.90$ to $1.65$, while near-baseline circuits see
marginal changes). AutoDMP ($1.20\times$ WNS) and MaskRegulate
($1.01\times$ WNS) improve timing through configuration tuning and RL,
though MaskRegulate exhibits high variance (ariane133 $0.60\times$ vs.\
ariane136 TNS $3.42\times$).
Re$^2$MaP ($0.89\times$ WNS, $0.61\times$ TNS) is the strongest
prior method, with particularly notable TNS on black\_parrot
($0.02\times$) and bp\_be ($0.51\times$).

PPAPlace-Refine alone applies gradient descent to DREAMPlace's default
placement, reducing average WNS from $1.61$ to $1.30$ and TNS from
$1.92$ to $1.36$. This improves over untuned DREAMPlace but the
local nature of gradient refinement limits gains on circuits where the
starting topology is already misaligned (ariane133 WNS remains
$2.15\times$).

PPAPlace-CoOpt injects the surrogate co-objective into DREAMPlace's
loop, achieving average WNS of $0.83$ and TNS of $0.57$. The
gap over DREAMPlace\,4.0 demonstrates the value of post-route
supervision over pre-route STA. Post-DRT reports confirm zero design-rule-check
violations and routed wirelength within $3\%$ of default, so the
co-objective does not degrade routability.

\begin{figure}[h]
    \centering
    \includegraphics[width=\columnwidth]{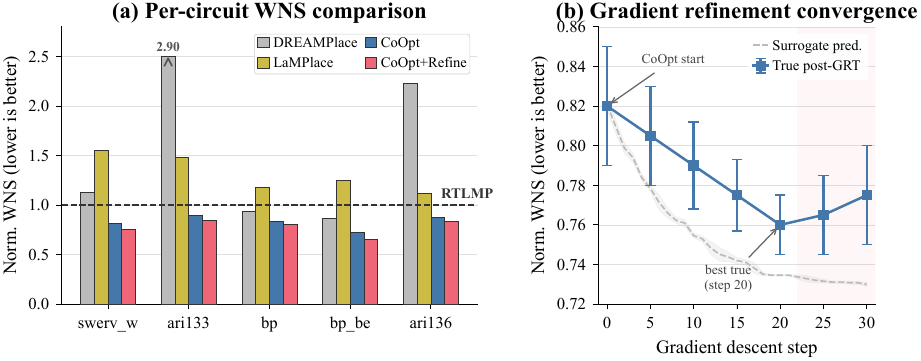}
    \caption{(a)~Per-circuit normalized WNS (Table~\ref{tab:main}).
    (b)~True vs.\ surrogate WNS during gradient refinement on swerv\_wrapper.}
    \label{fig:main}
\end{figure}

CoOpt{+}Refine's WNS/TNS ratios (improvements) are
$0.74/0.41$ ($26\%/59\%$) for in-family circuits, $0.85/0.62$
($15\%/38\%$) for out-of-family Ariane circuits, and
$0.78/0.49$ ($22\%/51\%$) overall.
This surpasses Re$^2$MaP's algorithmic approach ($0.89$/$0.61$) on both
metrics, demonstrating that a learned post-route objective
can outperform hand-engineered placement heuristics.
Unlike MaskRegulate's circuit-specific gains, CoOpt{+}Refine improves
consistently across all test circuits
(Figure~\ref{fig:main}(a)). It achieves the best per-circuit result
on 6 of 10 circuit--metric pairs and the best average on both WNS
and TNS. On the remaining four pairs, it yields the second-best result
except for the TNS of black\_parrot. The power ratios ($0.99$--$1.02\times$) confirm that PPAPlace's timing improvements do not increase power. On the same hardware, standard DREAMPlace takes about 24 seconds per placement, while CoOpt completes in under one minute, giving a placement-time overhead below $2.5\times$.

All the results reported in Table~\ref{tab:main} are final post-DRT PPA metrics. Although the surrogate is trained on post-GRT labels, the improvements carry
through to full detailed routing. This is consistent with the fidelity
analysis in Section~\ref{sec:label_fidelity}, where post-GRT rankings
preserve post-DRT outcomes with $\bar{\rho}=0.86$.

\subsection{Gradient Quality}
\label{subsec:gradient_quality}

Gradient-guided placement requires directionally accurate gradients.
\texttt{torch.autograd.gradcheck} confirms numerical correctness
(relative error $< 10^{-6}$) on 30 test placements. To assess
directional alignment, we perturb converged placements along 100
random directions per circuit and measure true PPA changes via
post-GRT. Table~\ref{tab:gradient_quality} reports cosine similarity
between surrogate and true gradients: average $0.53$ for WNS and
$0.46$ for TNS, positive on all circuits.

\begin{table}[h]
\centering
\caption{Cosine similarity between surrogate gradient and true post-GRT PPA change (100 perturbation directions per circuit).}
\label{tab:gradient_quality}
\renewcommand{\arraystretch}{1.15}
\footnotesize
\setlength{\tabcolsep}{3pt}
\begin{tabular}{@{}l cccccc@{}}
\toprule
 & swerv & ari133 & black & bp\_be & ari136 & Avg. \\
\midrule
cos($\nabla$\,WNS) & 0.49 & 0.58 & 0.51 & 0.62 & 0.43 & 0.53 \\
cos($\nabla$\,TNS) & 0.43 & 0.49 & 0.46 & 0.53 & 0.37 & 0.46 \\
\bottomrule
\end{tabular}
\setlength{\tabcolsep}{6pt}
\renewcommand{\arraystretch}{1.0}
\end{table}

Figure~\ref{fig:main}(b) validates this on swerv\_wrapper: gradient
descent reduces true post-GRT WNS from $0.82$ to $0.76$ over 20
steps, closely tracked by the surrogate. Beyond step~22, true WNS
rises as the placement exits the training distribution. The
refinement loop returns the checkpoint with the lowest surrogate loss
across all $T$ steps, mitigating moderate overshoot.

We sweep $\lambda_p \in \{0.001, 0.005, 0.01, 0.05\}$ on
swerv\_wrapper, yielding WNS of $\{0.92, 0.87, 0.82, 0.90\}$;
$\lambda_p = 0.01$ provides the best trade-off and is reused
unchanged across all test circuits.
For refinement steps, $T \in \{10, 20, 30, 50\}$ gives WNS
$\{0.79, 0.77, 0.76, 0.77\}$, with $T{=}30$ best and
$T{=}50$ slightly worse.
The learning rate $\alpha{=}0.001$ is selected by grid search.

\subsection{Generalization}
\label{subsec:generalization}

{
Table~\ref{tab:generalization} evaluates 500 held-out DREAMPlace
placements per circuit. Average WNS Spearman $\rho$ is $0.77$, Kendall
$\tau$ is $0.58$, and top-5 accuracy is 68\% versus 1\% for random
selection. Accuracy ranges from $\rho=0.72$ on ariane136 to $0.83$ on
bp\_be. Three test circuits share training-circuit lineages
(bp\_be/\allowbreak bp\_be12, swerv\_wrapper/\allowbreak swerv\_wrapper43,
black\_parrot/\allowbreak bp\_fe).
Shared design lineages can make these three cases easier
than unseen design families. The Ariane circuits, both
absent from training, reach $\rho=0.81$ and $0.72$, respectively, with
ariane136 lowest.
}

\begin{table}[h]
\centering
\caption{Surrogate generalization: ranking accuracy on held-out placements and cross-placer transfer to RTLMP.}
\label{tab:generalization}
\renewcommand{\arraystretch}{1.15}
\footnotesize
\setlength{\tabcolsep}{2pt}
\begin{tabular}{@{}l cccccccc@{}}
\toprule
Metric & swerv & ari133 & black & bp\_be & ari136
  & bp\_fe\textsuperscript{t} & ether\textsuperscript{t} & Avg. \\
\midrule
\multicolumn{9}{@{}l}{\itshape Held-out (DREAMPlace)} \\
$\rho$ & 0.74 & 0.81 & 0.77 & 0.83 & 0.72 & N/A & N/A & 0.77 \\
$\tau$ & 0.56 & 0.62 & 0.55 & 0.65 & 0.51 & N/A & N/A & 0.58 \\
Top-5 & 60\% & 80\% & 60\% & 80\% & 60\% & N/A & N/A & 68\% \\
\addlinespace[1pt]
\multicolumn{9}{@{}l}{\itshape Cross-placer (RTLMP)} \\
$\rho_\text{WNS}$ & 0.63 & 0.58 & 0.60 & 0.72 & 0.54 & 0.68 & 0.55 & 0.61 \\
$\rho_\text{TNS}$ & 0.57 & 0.54 & 0.53 & 0.66 & 0.50 & 0.60 & 0.52 & 0.56 \\
\bottomrule
\multicolumn{9}{@{}l}{\scriptsize\textsuperscript{t}Training circuit (cross-placer only).}
\end{tabular}
\setlength{\tabcolsep}{6pt}
\renewcommand{\arraystretch}{1.0}
\end{table}

\begin{figure}[h]
    \centering
    \includegraphics[width=\columnwidth]{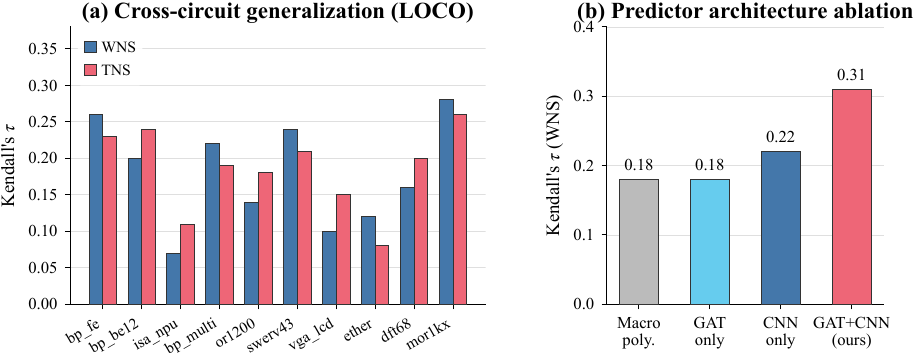}
    \caption{(a)~Leave-one-circuit-out (LOCO) cross-circuit generalization (Kendall's $\tau$).
    (b)~Predictor architecture ablation (Kendall's $\tau$).}
    \label{fig:analysis}
\end{figure}

Leave-one-circuit-out (LOCO) cross-validation
(Figure~\ref{fig:analysis}(a)) trains on 9 circuits and evaluates on the
tenth. Per-circuit $\tau$ ranges from $0.07$ on isa\_npu to $0.28$ on
mor1kx, which is sufficient to identify above-average placements.

The cross-placer rows of
Table~\ref{tab:generalization} apply the DREAMPlace-trained predictor to
unseen RTLMP placements. WNS $\rho$ is $0.61$ versus $0.77$ on
DREAMPlace and remains significant ($p<0.01$) on every circuit. This
drop indicates a placement-distribution gap rather than placer independence.

Table~\ref{tab:superblue} evaluates the same
checkpoint zero-shot on 100 IBM 45\,nm Superblue16/18 placements, 50
per circuit. WNS/TNS Spearman $\rho$ is $0.68/0.62$, with 56\% top-5
accuracy. Against LaMPlace~\cite{lamplace2025}, PPAPlace ranks second
in WNS, first on superblue16 TNS, and within 3.2\% on superblue18 TNS.

\begin{table}[h]
\centering
\caption{Final timing on Superblue16/18, with baselines from LaMPlace~\cite{lamplace2025}. WNS is in $10^3$\,ps and TNS in $10^5$\,ps. Values closer to zero are better.}
\label{tab:superblue}
\renewcommand{\arraystretch}{1.15}
\footnotesize
\setlength{\tabcolsep}{3pt}
{
\begin{tabular}{@{}lrrrr@{}}
\toprule
& \multicolumn{2}{c}{superblue16}
& \multicolumn{2}{c}{superblue18} \\
\cmidrule(lr){2-3}\cmidrule(lr){4-5}
Method & WNS & TNS & WNS & TNS \\
\midrule
DREAMPlace~\cite{lin2019dreamplace} & $-107.05$ & $-1526.10$ & $-88.11$ & $-751.27$ \\
WireMask-EA~\cite{shi2023wire} & $-635.89$ & $-18343.30$ & $-78.25$ & $-406.01$ \\
ChiPFormer~\cite{lai2023chipformer} & $-322.05$ & $-15426.07$ & $-80.57$ & $-378.90$ \\
LaMPlace~\cite{lamplace2025} & $-36.87$ & $-1514.73$ & $-66.93$ & $-426.91$ \\
PPAPlace zero-shot & $-68.50$ & $-1380.50$ & $-73.40$ & $-440.50$ \\
\bottomrule
\end{tabular}
}
\setlength{\tabcolsep}{6pt}
\renewcommand{\arraystretch}{1.0}
\end{table}

\subsection{Ablation Studies}
\label{subsec:ablations}

\subsubsection{Label Fidelity and Representation}
\label{subsec:2x2}

{
The upper section of Table~\ref{tab:ablation_combined} crosses two
label stages with two representations under identical training.
LaMPlace's macro-only + pre-route STA setting has
weak ranking agreement ($\tau=0.08$). Both axes contribute independently.
Post-GRT doubles macro-only $\tau$ from $0.08$ to $0.18$, GAT+CNN
raises pre-route $\tau$ from $0.08$ to $0.13$, and combining both
reaches $0.31$.
}

\begin{table}[h]
\centering
\caption{Ablation on labels and representation (Kendall's $\tau$, WNS). Upper: two label stages $\times$ two representations. Lower: label stages with GAT+CNN fixed.}
\label{tab:ablation_combined}
\renewcommand{\arraystretch}{1.15}
\footnotesize
\begin{tabular}{@{}ll cc@{}}
\toprule
Training labels & Architecture & $\tau$ (WNS) & Top-1 \\
\midrule
\multicolumn{4}{@{}l}{\emph{Label stage $\times$ representation}} \\[1pt]
Pre-route STA & Macro-only poly.\ & 0.08\tpm.02 & n/a \\
Pre-route STA & GAT+CNN & 0.13\tpm.03 & 26\% \\
Post-GRT      & Macro-only poly.\ & 0.18\tpm.02 & n/a \\
Post-GRT      & GAT+CNN & \textbf{0.31\tpm.02} & \textbf{52\%} \\
\midrule
\multicolumn{4}{@{}l}{\emph{Label stage (GAT+CNN fixed)}} \\[1pt]
Post-CTS      & GAT+CNN & 0.16\tpm.03 & 30\% \\
Post-DRT      & GAT+CNN & 0.34\tpm.02 & 56\% \\
\bottomrule
\end{tabular}
\renewcommand{\arraystretch}{1.0}
\end{table}

{
With GAT+CNN fixed, $\tau$ rises from $0.13$ with pre-route STA to
$0.16$ with post-CTS and $0.31$ with post-GRT, isolating the supervision
stage. Post-DRT reaches $0.34$ but costs $3.7$ hours per label versus
$0.20$ hours for post-GRT, which supports more training data under the
same offline budget.
}

\subsubsection{Predictor Architecture}
\label{subsec:pred_results}

Figure~\ref{fig:analysis}(b) compares four
post-GRT variants: macro-only polynomial ($\tau=0.18$), CNN-only
($0.22$), GAT-only ($0.18$), and GAT+CNN ($0.31$). The combined model
outperforms either stream alone, showing that density patterns and
netlist connectivity provide complementary information.


\section{Conclusion}
\label{sec:conclusion}

PPAPlace demonstrates that a differentiable surrogate trained on
post-GRT labels can provide gradient-based PPA feedback that no
analytical proxy achieves. CoOpt{+}Refine improves average WNS and
TNS by 22\% and 51\% over Hier-RTLMP on five held-out
circuits.

Despite the performance gains, several limitations suggest promising future directions. Training uses Nangate45, while the zero-shot Superblue test uses the IBM 45\,nm library. Broader validation
across technology nodes and standard-cell libraries remains a natural
next step. The raw refinement trajectory exhibits
out-of-distribution degradation after about 20 steps
(Figure~\ref{fig:main}(b)).
Distribution-aware stopping or trust-region constraints could improve
robustness. Cross-placer transfer ($\rho = 0.61$) lags behind
within-placer accuracy ($0.77$). Targeted fine-tuning or
domain-adaptation techniques could narrow this gap. Finally, CoOpt
currently requires integration with a differentiable placer.
Extending the co-objective paradigm to commercial tools that expose
only final placements remains an open challenge that the
input-compatible Refine mode partially addresses.


\begin{acks}
Supported by the
\grantsponsor{NSERC}
{Natural Sciences and Engineering Research Council of Canada (NSERC)}
{https://www.nserc-crsng.gc.ca/}
(\grantnum{NSERC}{RES0048688},
\grantnum{NSERC}{RES0051374}, and
\grantnum{NSERC}{RES0054326}) and
\grantsponsor{AlbertaInnovates}
{Alberta Innovates}
{https://albertainnovates.ca/}
(\grantnum{AlbertaInnovates}{RES0053965}).
\end{acks}

\bibliographystyle{ACM-Reference-Format}
\bibliography{references}

\end{document}